\documentclass[sigconf]{acmart}
\AtBeginDocument{%
  }

\setcopyright{acmlicensed}
\copyrightyear{2018}
\acmYear{2018}
\acmDOI{XXXXXXX.XXXXXXX}
\acmConference[Conference acronym 'XX]{Make sure to enter the correct
  conference title from your rights confirmation email}{June 03--05,
  2018}{Woodstock, NY}
\acmISBN{978-1-4503-XXXX-X/2018/06}

\usepackage{amsmath}
\usepackage[utf8]{inputenc} % allow utf-8 input
\usepackage[T1]{fontenc}    % use 8-bit T1 fonts
\usepackage{hyperref}       % hyperlinks
\usepackage{url}            % simple URL typesetting
\usepackage{booktabs}       % professional-quality tables
\usepackage{amsfonts}       % blackboard math symbols
\usepackage{nicefrac}       % compact symbols for 1/2, etc.
\usepackage{microtype}      % microtypography
\usepackage{xcolor}         % colors
\usepackage{multirow}
\usepackage{subfigure,comment,adjustbox}
\usepackage{makecell}

\usepackage{subcaption}

\usepackage{xspace}
\usepackage{graphicx}
\newcommand{\dataset}{\textsc{ReceiptSense}\xspace}
\begin{document}

\title{ReceiptSense: Beyond Traditional OCR - A Dataset for Receipt Understanding}

\author{Abdelrahman Abdallah}
\affiliation{%
  \institution{ Innsbruck University}
  \city{Innsbruck}
  %\postcode{28644}
  \country{Austria}
}
\email{abdelrahman.abdallah@uibk.ac.at}

\author{Mohamed Mounis}
\affiliation{%
  \institution{ High Institute for Computers and Management Information }
  \city{Alexandria}
  %\postcode{28644}
  \country{Egypt}
}
\email{mahmoudelsayed@chungbuk.ac.kr}

\author{Mahmoud Abdalla}
\affiliation{%
  \institution{ Chungbuk National University}
  \city{Cheongju}
  \postcode{28644}
  \country{Republic of Korea}
}
\email{mahmoudelsayed@chungbuk.ac.kr}

\author{Mahmoud SalahEldin Kasem}
\affiliation{%
\institution{ Chungbuk National University}
  \city{Cheongju}
  %\postcode{28644}
  \country{Republic of Korea}
}
\email{mahmoud.kasem@cbnu.ac.kr}

\author{Mohamed Mahmoud}
\affiliation{%
  \institution{ Chungbuk National University}
  \city{Cheongju}
  %\postcode{28644}
  \country{Republic of Korea}
}
\email{mohamedabokhalil@aun.edu.eg}
\author{
 Ibrahim Abdelhalim}
\affiliation{%
  \institution{Louisville University}
  \city{Louisville}
  %\postcode{28644}
  \country{USA}
}
\email{ibrahim.abdelhalim@louisville.edu}

\author{ Mohamed Elkasaby}
\affiliation{%
  \institution{DISCO}
  \city{Cairo}
 \state{Cairo}
  \country{Egypt}
  }
\email{mo.ab.elkasaby@gmail.com	}

\author{Yasser	Elbendary}
\affiliation{%
  \institution{DISCO}
  \city{Cairo}
 \state{Cairo}
  \country{Egypt}
  }
\email{yelbendary@discoapp.ai	}

\author{Adam Jatowt}
%\orcid{0000-0001-7235-0665}
\affiliation{%
  \institution{University of Innsbruck}
  \city{Innsbruck}
 \state{Tyrol}
  \country{Austria}
  }
\email{adam.jatowt@uibk.ac.at}



\renewcommand{\shortauthors}{Abdelrahman et al.}

\begin{abstract}
Multilingual OCR and information extraction from receipts remains challenging, particularly for complex scripts like Arabic. We introduce \dataset, a comprehensive dataset designed for Arabic-English receipt understanding comprising 20,000 annotated receipts from diverse retail settings, 30,000 OCR-annotated images, and 10,000 item-level annotations, and a new Receipt QA subset with 1265 receipt images paired with 40 question-answer pairs each to support LLM evaluation for receipt understanding. The dataset captures merchant names, item descriptions, prices, receipt numbers, and dates to support object detection, OCR, and information extraction tasks. We establish baseline performance using traditional methods (Tesseract OCR) and advanced neural networks, demonstrating the dataset's effectiveness for processing complex, noisy real-world receipt layouts. Our publicly accessible dataset advances automated multilingual document processing research\footnote{\url{https://github.com/Update-For-Integrated-Business-AI/CORU}}.
\end{abstract}





\maketitle

\section{Introduction}
\begin{figure}[h]
    \centering		
    %\subfigure[]{
        \includegraphics[width=0.5\textwidth,height=0.40\textwidth]{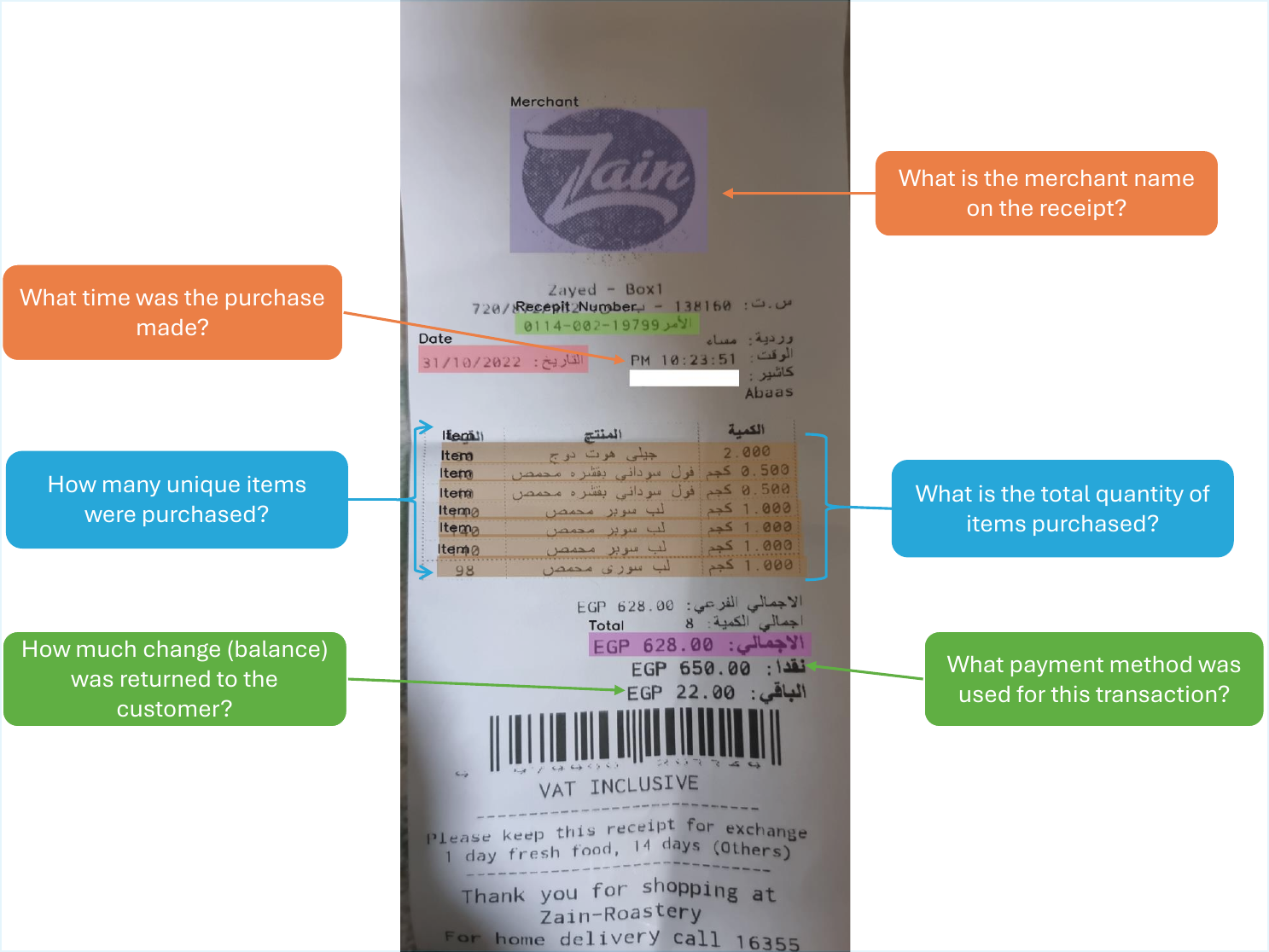}
        %\label{fig:class_distribution}
   % }
   % \subfigure[]{
        %\includegraphics[width=0.20\textwidth,height=0.40\textwidth]{figures/1b5f7808-9ebe-4538-b962-8dc3c26c14f6.jpg}
        %\label{fig:language}
   % }
    \caption{ Examples of annotated receipt images from the \dataset dataset.% The images illustrate the variety of receipt formats and the complexity of text layouts.
    }
    \label{fig:example_images}	 		
\end{figure}

Optical Character Recognition (OCR) \cite{Nguyen_Thi,hwang2019post,nguyen2021mc} converts images of characters into digital text. While deep learning has improved OCR performance, challenges remain in post-OCR parsing—predicting semantic labels from noisy, unstructured OCR output, especially for receipts with diverse layouts, fonts, and degraded quality. Existing datasets fall short: standard OCR datasets lack parsing labels, while parsing datasets contain clean text that does not reflect real OCR errors. Recent efforts like SROIE~\cite{huang2019icdar2019} and CORD~\cite{park2019cord} offer annotated receipts but do not comprehensively address multilingual understanding. We introduce \textbf{\dataset}, the first public dataset for advanced multilingual post-OCR parsing, featuring diverse Arabic-English receipts with detailed OCR, key information, and item-level annotations. Its four components—text-level annotations, key information labels, and item-level data—support OCR enhancement, information extraction, QA  and multilingual text analysis. Figure~\ref{fig:example_images} shows examples highlighting the variety of formats and complex layouts.

The primary contributions of the \dataset dataset are as follows: (1) \textbf{Key Information Detection:} 20,000 human-annotated receipts extracting merchant names, dates, receipt numbers, item lists, and totals for key-value information extraction from noisy text. (2) \textbf{Large-Scale OCR Dataset:} 30,000 receipt images with box-level text annotations addressing challenges from diverse layouts, fonts, and noisy appearances.
(3) \textbf{Detailed Item Analysis:} 10,000 individual items annotated for item-level analysis in expense management, inventory tracking, and retail analytics. 
(4) \textbf{Receipt QA:} includes 1,265 receipt images, where each is paired with 40 question-answer pairs covering merchant, date, totals, items, taxes, and payment methods. 
(5) \textbf{Baseline Performance Metrics:} Comprehensive benchmarks using traditional and modern deep learning approaches across object detection, OCR, and information extraction tasks.

\dataset introduces several novel contributions compared to existing datasets (SROIE, CORD, UIT, MC-OCR). It is significantly larger with \textbf{20,000 annotated receipts}, \textbf{50,600 QA}, \textbf{30,000 OCR-annotated images}, and \textbf{10,000 item-level annotations}. Unlike existing English-only datasets with clean text, \dataset incorporates \textbf{multilingual Arabic-English receipts} with diverse layouts, fonts, and mixed-language content reflecting global retail scenarios. The dataset includes \textbf{real-world noise}—blurred, rotated, or damaged text—challenging models for robust performance. Crucially, \dataset is the first to provide \textbf{item-level annotations}, enabling fine-grained extraction for \textbf{expense management} and \textbf{inventory analysis}. These characteristics advance multilingual OCR, post-OCR parsing, and practical applications in finance and retail.

\begin{figure}[h]
    \centering
    % First row with three subfigures
    \subfigure[]{
        \includegraphics[width=0.20\textwidth]{figures/ocr/00b61ef0-fb11-44a0-9629-2eb120844689_date.jpg}
    }
    \subfigure[]{
        \includegraphics[width=0.2\textwidth]{figures/ocr/00b7f1ac-a055-4b79-b5f8-54d7f65e61aa_merchant.jpg}
    }
    \subfigure[]{
        \includegraphics[width=0.2\textwidth, height=0.4cm]{figures/ocr/00bc1fff-5438-4d7e-a40f-9714f9acbf7f_merchant.jpg}
    }
    \subfigure[]{
        \includegraphics[width=0.2\textwidth]{figures/ocr/0a295b55-5c3d-4f4e-830d-fc3b2edb63d8_line_12.jpg}
    }
    % Line break for new row
    \\
    % Second row with two subfigures
    \subfigure[]{
        \includegraphics[width=0.2\textwidth]{figures/ocr/00b7f1ac-a055-4b79-b5f8-54d7f65e61aa_total.jpg}
    }
    \subfigure[]{
        \includegraphics[width=0.2\textwidth]{figures/ocr/00b61ef0-fb11-44a0-9629-2eb120844689_line_2.jpg}
    }

    \caption{OCR images from receipt. Each subfigure shows a different example of the OCR task, including date, merchant, and items.}
    \label{fig:ocr_images}
\end{figure}

\section{Related Work}
While several datasets cater to various OCR tasks~\cite{Nguyen_Thi,abdallah2024transformers}, there is a notable shortage of receipt datasets, particularly for Arabic receipts. To highlight the unique features and comprehensive nature of our dataset, Table \ref{tab:exitsing_datsets} compares it against existing datasets like SROIE, MC-OCR, UIT, and CORD.

One of the earliest and most popular datasets in the scanned receipt domain is the ICDAR 2019 Challenge on Scanned Receipts OCR and Key Information Extraction (SROIE) dataset \cite{huang2019icdar2019}. This dataset marked a significant advancement in the automated analysis of scanned receipts by introducing a standardized collection of 1,000 annotated receipt images. It focuses on three tasks: text localization, OCR, and key information extraction. These tasks are essential for document analysis systems with substantial commercial potential. The challenge emphasized the unique difficulties inherent in receipt OCR, such as poor scan quality and complex layouts. SROIE has since become a cornerstone for research in receipt document understanding, serving as a reliable benchmark for evaluating the performance of novel models in various OCR and information extraction tasks.

The CORD (Consolidated Receipt Dataset) dataset \cite{park2019cord} addresses the challenge of integrating OCR with NLP tasks like semantic parsing by providing a comprehensive resource for post-OCR parsing. It includes thousands of Indonesian receipt images with box-level text annotations for OCR and multi-level semantic labels for parsing. Unlike traditional datasets, CORD bridges the gap between OCR and parsing, enabling the development of robust models that can handle OCR errors. It also introduces line annotations for converting two-dimensional OCR text into a well-ordered sequence, thereby enhancing parsing performance. CORD's hierarchical labeling structure allows researchers to investigate both low-level text extraction and high-level semantic interpretation, making it a versatile resource for document understanding tasks.

The MC-OCR (Mobile Captured Receipt Recognition Challenge) dataset \cite{vu2021mc}, featured at RIVF 2021 conference, includes 2,436 images of Vietnamese receipts captured via mobile devices. This dataset supports two tasks: predicting receipt quality and recognizing key information fields. Collected by 50 data collectors, these images were annotated in two phases: image quality assessment (IQA) and key information extraction (KIE). IQA involved evaluating text line readability, while KIE required annotators to identify and transcribe key fields. This dataset serves as a benchmark for improving document digitalization and automated financial document processing, with evaluation metrics including RMSE for IQA and CER for KIE. Its mobile-captured nature makes it particularly relevant to real-world applications where varying lighting conditions and image distortions are common challenges.

The UIT-MLReceipts dataset \cite{nguyen2024uit} addresses the need for a more extensive and carefully labeled dataset for extracting receipt information, overcoming the limitations of existing datasets like SROIE and CORD. With a focus on enriching the available data, UIT-MLReceipts has been compiled by sourcing receipts from various establishments such as restaurants, cafes, bookstores, and supermarkets, ensuring diversity in structure, color, font, and format. Additionally, images from social media groups were incorporated to further enhance dataset variability. Following meticulous curation, a total of 2,147 receipt images were obtained and annotated, covering key information like store names, addresses, timestamps, and total costs. These annotations were refined using a Faster R-CNN model trained on the MC-OCR challenge dataset, ensuring accuracy and completeness. The resulting UIT-MLReceipts dataset offers a comprehensive representation of receipt characteristics, languages, status, and image attributes, making it a valuable asset for research in receipt information extraction and Visual Document Understanding (VDU).

Despite the existence of these datasets, there remains a substantial gap in resources tailored for Arabic receipt information extraction. Arabic receipts often present unique challenges such as right-to-left text orientation, diverse numeral systems, and variations in layout structures. Our proposed CORU dataset aims to address these challenges by offering a sizable, high-quality collection of Arabic and English receipts annotated with key information fields. This dataset not only facilitates OCR tasks but also supports advanced question-answering models by including context-aware questions related to receipt content. By bridging this gap, CORU contributes to the broader effort of developing multilingual and multimodal OCR systems capable of handling diverse document structures across languages and regions.

\begin{table*}[h]
\small
\centering
\caption{Comparative overview of the \dataset dataset, highlighting differences in the number of images, categories, and supported tasks. OB refers to Object detection and IE refers to Information Extraction. }
\label{tab:exitsing_datsets}
\begin{adjustbox}{width=0.8\textwidth}
\begin{tabular}{c|c|c|c|c|c|c|c|c} %|c
\hline
\textbf{Dataset Name} & \textbf{\# Images} &  \textbf{\# categories} &\textbf{OB} & \textbf{OCR} & \textbf{IE} & \textbf{Item IE}  & \textbf{ReceiptSense} & \textbf{Language}   \\ %& \textbf{Year}
\hline
SROIE & 1,000 & 4  &\checkmark & \checkmark & \checkmark &\checkmark &  $X$&English \\ %& 2019 
\hline

MC-OCR & 2,436 & 4  & $X$ & \checkmark & \checkmark &\checkmark &  $X$&English \& Vietnamese \\ %& 2021
\hline
UIT & 2,147 & 4  & \checkmark & \checkmark & \checkmark &\checkmark & $X$ &English \& Vietnamese \\ %& 2022
\hline
CORD & 1,000 & 8 & \checkmark & \checkmark  &\checkmark & \checkmark &  $X$&English\\ % & 2019

\hline
\textbf{\dataset (ours)} & \textbf{20,000} &  \textbf{5}  & \textbf{\checkmark} & \textbf{\checkmark} & \textbf{\checkmark}  &\textbf{\checkmark} & \textbf{\checkmark} &\textbf{English \& Arabic} \\ %& 2024
\hline
\end{tabular}
\end{adjustbox}
\end{table*}
\section{Dataset Creation and Analysis}

\subsection{Data Collection }
We developed \dataset to advance multilingual receipt understanding, particularly for Arabic and English texts. Our methodology integrates rigorous data collection, annotation, and quality control processes to ensure diversity and real-world applicability. \textbf{Ethical Considerations and Data Collection:}
All receipts were collected with explicit user consent through the DISCO application\footnote{\url{https://discoapp.ai/}}, following strict privacy protocols. We gathered over 100k receipts from restaurants, supermarkets, and retail stores across different geographical regions. To protect privacy, our annotation team applied a four-step PII redaction process: line-by-line review, sensitive data obscuring, verification, and independent cross-checking. \textbf{Annotation Process: } We established detailed guidelines for annotating merchant names, items, prices, and dates, with specialized protocols for bi-directional Arabic-English text. Using MakeSense\footnote{\url{https://www.makesense.ai/}}, annotators created bounding boxes in YOLO and COCO formats. For OCR tasks, we developed a custom system maintaining positional integrity of mixed-language text. Item-specific annotations covered names, classifications, quantities, prices, packaging, and brands, validated through iterative feedback loops with domain experts. For the Receipt QA subset, we used 1,265 real receipt images and manually authored 40 diverse QA pairs per receipt. Questions cover receipt meta-data, item-level details, and transaction summaries. The QA pairs were validated for consistency with the receipt content.

\subsection{Dataset  Statistical Analysis}
\dataset demonstrates significant diversity across multiple dimensions. The language distribution shows Arabic predominance (53.6\%), followed by English (26.2\%) and mixed-language content (20.3\%), reflecting real-world multilingual scenarios in international retail environments. Object class analysis reveals \texttt{Item} entries as most frequent, supporting detailed transaction analysis for applications like itemised billing and inventory management. The item class distribution spans from prevalent categories like 'Soft drinks' and 'Rice, pasta, and noodles' to specialised items like 'Hair \& body care', ensuring model versatility across commercial domains.  The Receipt QA subset comprises over 50,000 QA pairs (1,265 receipts × 40 questions). The QA coverage includes 30\% merchant/payment/date metadata, 50\% item-level information, and 20\% tax/total/payment details, ensuring comprehensive evaluation of receipt understanding.

\section{Models}

%We evaluate various deep learning models on \dataset, spanning traditional approaches like Tesseract OCR~\cite{patel2012optical} to advanced neural architectures. Our evaluation covers CNNs~\cite{o2015introduction}, Transformers, and cutting-edge language models across object detection, OCR, and information extraction tasks.

\paragraph{Visual QA Models}
To evaluate receipt-level question answering, we applied state-of-the-art large language models (LLMs) including GPT-4o~\cite{achiam2023gpt}, Llama 3.2~\cite{touvron2023llama2}, Phi3~\cite{abdin2024phi}, Phi3.5~\cite{abdin2024phi}, Llava~\cite{liu2023visual}, Internvl2 4B~\cite{chen2024expanding}, and Internvl2 8B~\cite{chen2024expanding}. These models were tasked with predicting answers to 40 distinct question types per receipt, covering receipt metadata, item details, VAT, totals, and payment methods.% We evaluated models across precision, recall, F1 score, exact match, and contains metrics.

\paragraph{Object Detection Models:} Weakly supervised models localize objects using only image-level labels, reducing annotation effort while maintaining accuracy~\cite{zhang2021weakly,choe2020evaluating}. Techniques include Class Activation Mapping (CAM)~\cite{zhou2016learning}, Hide-and-Seek (HAS)~\cite{singh-iccv2017}, Adversarial Complementary Learning (ACoL)~\cite{zhang2018adversarial}, Self-produced Guidance (SPG)~\cite{zhang2018self}, Attention-based Dropout Layer (ADL)~\cite{choe2019attention}, and CutMix~\cite{yun2019cutmix}, each enhancing feature learning through masking, attention, or augmentation. Advanced architectures like DINO~\cite{zhang2022dino} use Transformer-based deformable attention for precise localization, while YOLO models (YOLOv7~\cite{wang2023yolov7}, YOLOv8~\cite{jocher2023yolo}) offer real-time detection with multi-scale and attention mechanisms.

\paragraph{OCR Models}
Our OCR model combines CNNs and bidirectional LSTMs for text recognition. The convolutional layer performs feature extraction through: $f(x, y) = \sum_{i=-a}^a \sum_{j=-b}^b k(i, j) \cdot g(x-i, y-j)$, where $f(x, y)$ is the output feature map, $g(x, y)$ the input image, and $k(i, j)$ the convolutional filter. LSTM units capture sequential dependencies for accurate text decoding, with bidirectional processing ensuring comprehensive context understanding.

\paragraph{Large Language Models}
We evaluate several LLMs for information extraction tasks. Llama~\cite{llama_v2}, Mistral~\cite{mistral}, Mixtral~\cite{mixtral}, Falcon~\cite{falcon} and Zephyr~\cite{zephyr}.

\begin{table}[t]
\caption{Comparative Analysis of Object Detection Models Across Different Backbone Architectures.} 
\label{tab:comparison_all_models}
\centering
\begin{adjustbox}{width=0.40\textwidth}
\begin{tabular}{@{}lllccccccccc@{}}
\toprule
\multirow{2}{*}{\textbf{Method}} & \multirow{2}{*}{\textbf{Backbone}} & \multirow{2}{*}{\textbf{Avg}}& \multicolumn{9}{c}{IoU} \\

& & &  \textbf{10} & \textbf{20} & \textbf{30} & \textbf{40} & \textbf{50} & \textbf{60} & \textbf{70} & \textbf{80} & \textbf{90} \\ 
\midrule
\multirow{3}{*}{CAM} 
 & ResNet50  & 6.17 & 41.08 & 11.93 & 4.48 & 2.24 & 1.07 & 0.59 & 0.21 & 0.05 & 0.05 \\
 & VGG16  & 5.86 & 45.23 & 10.12 & 2.24 & 0.64 & 0.16 & 0.11 & 0.05 & 0.00 & 0.00 \\
 & InceptionV2  & 5.26 & 36.81 & 9.06 & 3.52 & 1.76 & 0.80 & 0.37 & 0.21 & 0.05 & 0.00 \\
\midrule
\multirow{3}{*}{HAS} 
 & ResNet50  & 7.14 & 43.26 & 16.84 & 6.55 & 2.72 & 1.17 & 0.59 & 0.21 & 0.11 & 0.00 \\
 & VGG16  & 0.00 & 0.00 & 0.00 & 0.00 & 0.00 & 0.00 & 0.00 & 0.00 & 0.00 & 0.00 \\
 & InceptionV2  & 4.65 & 34.79 & 8.20 & 2.40 & 0.75 & 0.21 & 0.11 & 0.05 & 0.00 & 0.00 \\
\midrule
\multirow{3}{*}{ADL} 
 & ResNet50  & 6.57 & 39.74 & 15.13 & 6.34 & 2.56 & 1.17 & 0.48 & 0.16 & 0.11 & 0.05 \\
 & VGG16  & 6.97 & 47.52 & 15.34 & 4.69 & 1.60 & 0.37 & 0.16 & 0.05 & 0.00 & 0.00 \\
 & InceptionV2  & 7.04 & 52.10 & 11.61 & 4.00 & 1.60 & 0.64 & 0.27 & 0.11 & 0.05 & 0.00 \\
\midrule
\multirow{3}{*}{ACOL} 
 & ResNet50  & 3.94 & 27.49 & 7.19 & 2.72 & 1.01 & 0.48 & 0.32 & 0.21 & 0.00 & 0.00 \\
 & VGG16  & 0.00 & 0.00 & 0.00 & 0.00 & 0.00 & 0.00 & 0.00 & 0.00 & 0.00 & 0.00 \\
 & InceptionV2  & 7.04 & 52.10 & 11.61 & 4.00 & 1.60 & 0.64 & 0.27 & 0.11 & 0.05 & 0.00 \\
\midrule
\multirow{3}{*}{SPG} 
 & ResNet50  & 6.53 & 42.14 & 13.00 & 5.06 & 2.66 & 1.33 & 0.64 & 0.27 & 0.11 & 0.05 \\
 & VGG16  & - & - & - & - & - & - & - & - & - & - \\
 & InceptionV2  & 4.74 & 35.11 & 7.73 & 2.66 & 1.07 & 0.48 & 0.27 & 0.05 & 0.00 & 0.00 \\
\midrule
\multirow{3}{*}{Cutmix} 
 & ResNet50  & 6.32 & 41.61 & 13.37 & 5.01 & 1.97 & 0.69 & 0.37 & 0.16 & 0.05 & 0.00 \\
 & VGG16  & 5.54 & 38.47 & 11.45 & 3.25 & 1.28 & 0.59 & 0.27 & 0.11 & 0.00 & 0.00 \\
 & InceptionV2  & 4.64 & 31.91 & 8.42 & 3.57 & 1.55 & 0.59 & 0.27 & 0.05 & 0.00 & 0.00 \\
\midrule
\multirow{3}{*}{DINO} 
 & Swin 4-scale & \textbf{32.2} & 45.4 & 44.6 & 43.3 & \textbf{41.9} & \textbf{39.9} & \textbf{35.9} & \textbf{27.5} & \textbf{10.6} & \textbf{0.7} \\
 & ResNet50 4-scale  & 31.9 & \textbf{45.9} & \textbf{45.0} & \textbf{43.6} & \textbf{41.9} & 39.4 & 35.2 & 25.6 & 10.2 & 0.5 \\
 & ResNet50 5-scale  & 29.4 & 44.1 & 43.2 & 41.7 & 39.8 & 37.1 & 32.5 & 25.0 & 0.89 & 0.00 \\
\bottomrule
\end{tabular}
\end{adjustbox}

\end{table}

\begin{figure}
    \centering
    \includegraphics[width=1\linewidth]{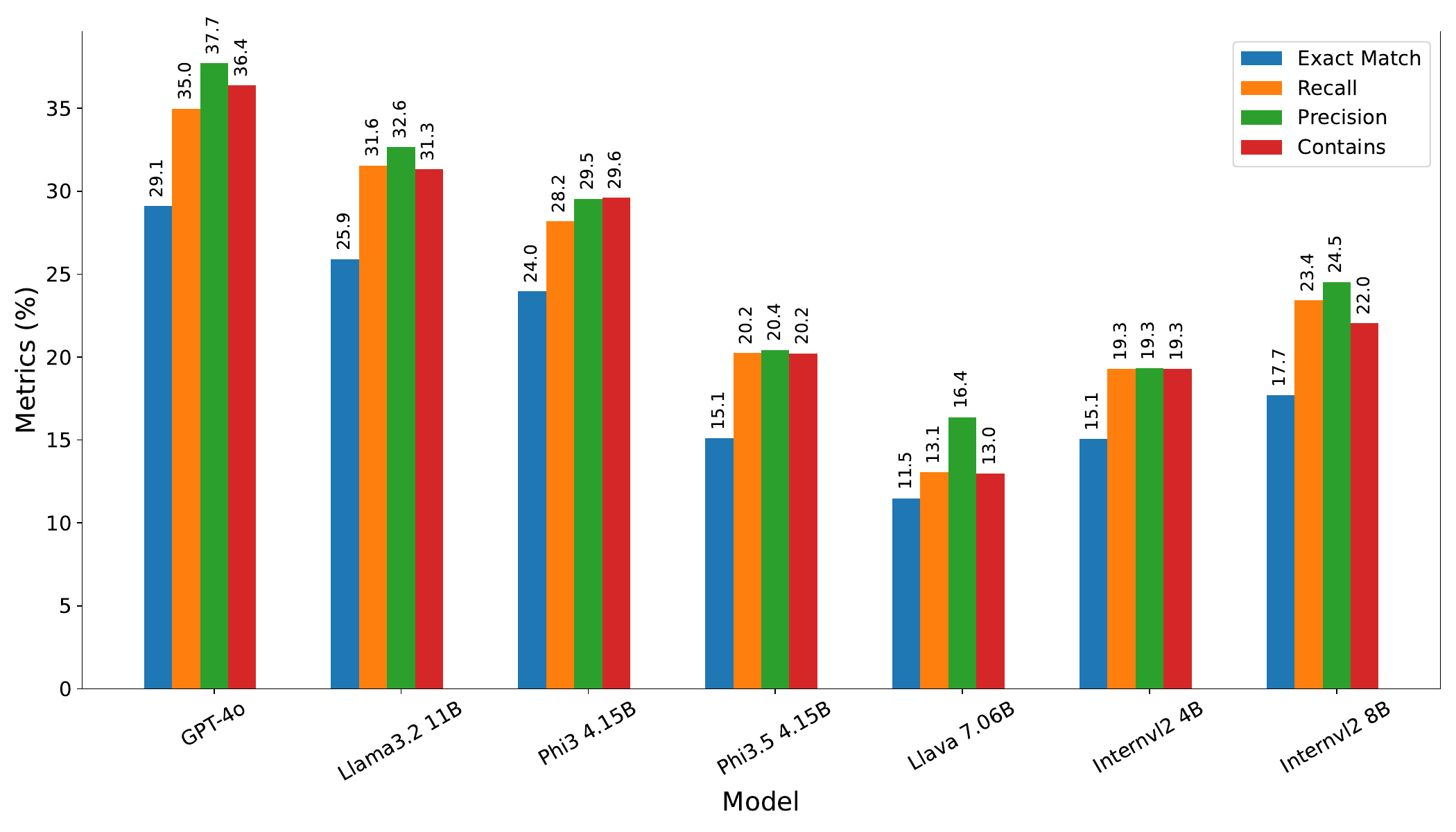}
    \caption{Performance of different models on the Receipt QA subset across key metrics.}
    \label{fig:evaluation_results_bar_chart_annotated}
\end{figure}

\begin{figure}
    \centering
    \includegraphics[width=.8\linewidth]{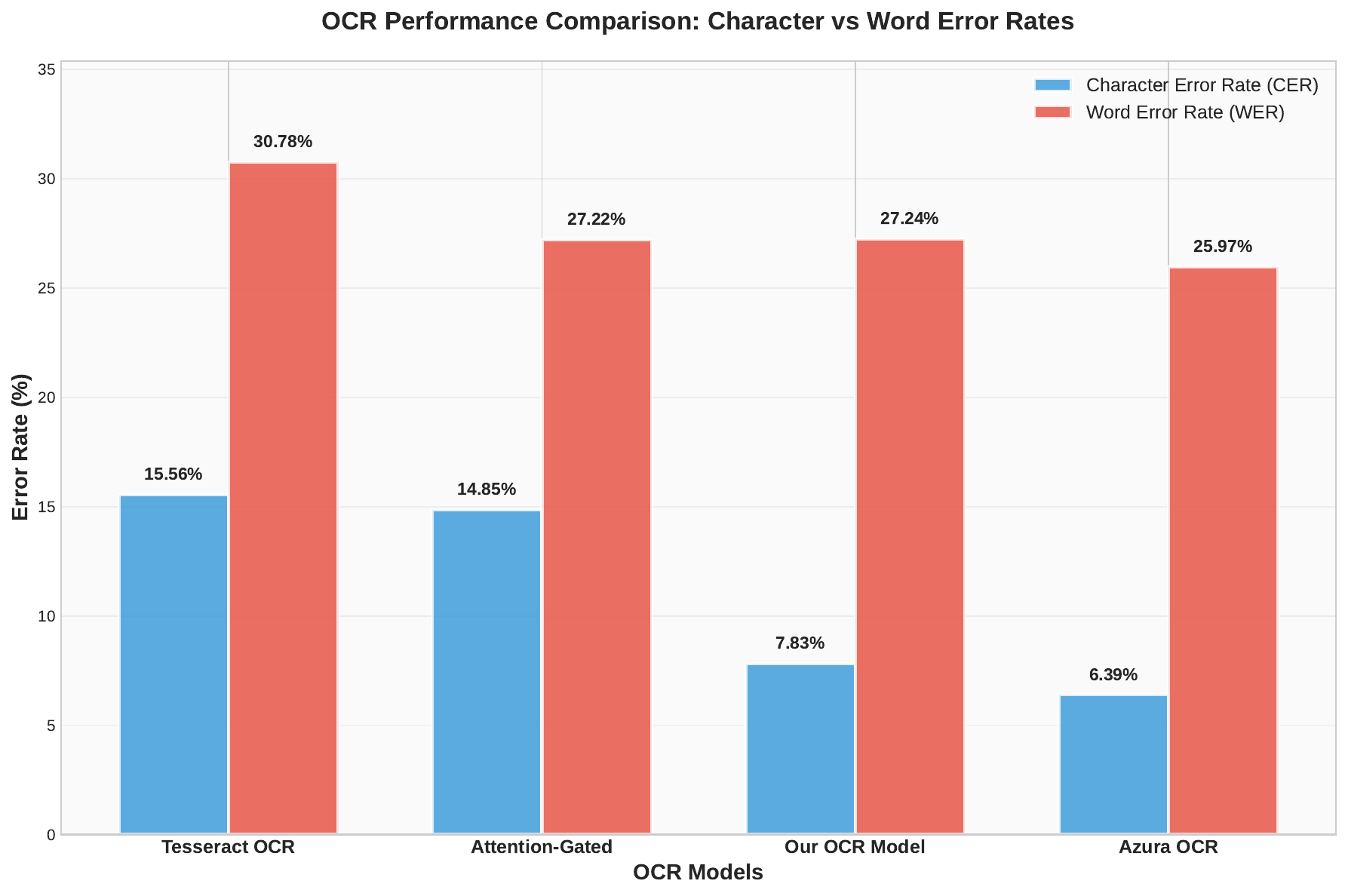}
    \caption{OCR Performance Comparison}
    \label{tab:ocr_performance_comparison}
\end{figure}

\begin{table}[h]
\begin{minipage}[t]{0.5\textwidth}
\centering
\caption{Performance Evaluation of YOLO Models}
\label{yolo_comparison}
\begin{tabular}{@{}lcccc@{}}
\toprule
Model & P & R & mAP50 & mAP50-95 \\
\midrule
YoloV7 & \textbf{76.00} & \textbf{85.60} & \textbf{79.20} & 43.70 \\
YoloV8 & 74.60 & 81.00 & 76.10 & 45.30 \\
YoloV9 & 75.70 & 83.40 & 77.90 & \textbf{46.70} \\
\bottomrule
\end{tabular}

\end{minipage}
\end{table}

\section{Experimental Results}
\subsection{QA Results}
Figure~\ref{fig:evaluation_results_bar_chart_annotated} provides a comparative analysis of various large language models on the Receipt QA subset of \dataset across four key metrics: precision, recall, exact match, and contains. GPT-4o consistently outperforms other models, achieving the highest scores across all metrics — notably 37.7\% precision, 36.4\% recall, 35.0\% exact match, and 29.1\% contains. Llama3.2 (11B) follows, with precision at 32.6\%, recall at 31.3\%, exact match at 31.6\%, and contains at 25.9\%. Phi3 (4.15B) and Phi3.5 show solid but slightly lower performance, clustering in the 28–30\% range for precision, recall, and exact match. Llava (7.06B) and Internvl2 models (4B, 8B) exhibit notably lower results, especially in the exact match and contains metrics, reflecting challenges in handling complex receipt QA tasks. The figure illustrates the clear advantage of larger and more advanced models like GPT-4o and Llama3.2 in comprehending and extracting accurate information from multilingual receipts.

\begin{table*}[t]
\caption{Performance of Language Models in Information Extraction Across Zero-Shot, One-Shot, Two-Shot, and Three-Shot Settings for Various Receipt Information Categories.% Metrics include F1 scores and accuracy for extracting \texttt{Brand}, \texttt{Weight}, \texttt{\#Units}, \texttt{Subtotal Units}, \texttt{Total Price}, \texttt{Price}, \texttt{Packaging}, and \texttt{Units} from multilingual receipts.
}

\centering
%\footnotesize % Reduced font size
%\setlength{\tabcolsep}{0.5mm}
\begin{adjustbox}{width=0.8\textwidth}
\begin{tabular}{c|c|c|cc|cc|cc|cc|cc|cc|cc|cc|cc@{}}
\toprule
\multirow{2}{*}{Models} & \multirow{2}{*}{parameters} & \multirow{2}{*}{\#Shots} & \multicolumn{2}{c|}{Brand} & \multicolumn{2}{c|}{Weight} & \multicolumn{2}{c|}{\# Units} & \multicolumn{2}{c|}{S.Units} & \multicolumn{2}{c|}{T.Price} & \multicolumn{2}{c|}{Price} & \multicolumn{2}{c|}{Pack} & \multicolumn{2}{c|}{Units} & \multicolumn{2}{c}{Overall}\\
& & & F1  & Acc  & F1  & Acc & F1 & Acc & F1  & Acc & F1 & Acc & F1  & Acc & F1  & Acc & F1  & Acc & F1  & Acc\\
\midrule
\textbf{LLaMA V1}& 7B & 0 & 3.70 & 0.93 & 4.29 & 1.25 & 8.14 & 3.34 & 0.04 & 0.02 & 0.08 & 0.04 & 0.08 & 0.04 & 0.08 & 0.04 & 0.00 & 0.00 & 0.88 & 0.46 \\
\textbf{LLaMA V2}& 7B & 0 & 29.16 & 16.46 & 28.32 & 15.87 & 74.94 & 60.52 & 0.04 & 0.02 & 0.75 & 0.38 & 1.50 & 0.75 & 0.33 & 0.17 & 2.39 & 0.24 & 20.70 & 10.80 \\
\textbf{LLaMA V2}& 13B & 0 & 14.46 & 6.97 & 11.70 & 5.35 & 54.00 & 36.89 & 0.00 & 0.00 & 0.79 & 0.40 & 0.67 & 0.34 & 0.17 & 0.08 & 7.19 & 2.82 & 12.14 & 5.61 \\
\textbf{Mistral}& 7B & 0 & 32.07 & 18.53 & 25.66 & 14.05 & 82.64 & \textbf{71.37} & 0.21 & 0.10 & \textbf{1.50} & \textbf{0.75} & 1.95 & 0.98 & 0.67 & 0.34 & 16.52 & 8.20 & 24.53 & 13.29 \\
\textbf{Mixtral}& 8x7B & 0 & 28.98 & 16.33 & 20.93 & 10.95 & 69.35 & 53.46 & 0.08 & 0.04 & 1.08 & 0.54 & \textbf{2.07} &\textbf{ 1.05} & 0.08 & 0.04 & 0.65 & 0.34 & 18.18 & 9.22 \\
\textbf{Falcon}& 7B & 0 & 29.88 & 16.96 & 18.60 & 9.48 & 48.80 & 32.05 & \textbf{0.25} & \textbf{0.13} & 0.88 & 0.44 & 0.96 & 0.48 & 0.88 & 0.44 & 0.00 & 0.00 & 13.24 & 6.25 \\
\textbf{Zephyr}& 7B & 0 & \textbf{42.98} & \textbf{27.02} & \textbf{33.89} & \textbf{19.87} & \textbf{77.88 }& 64.50 & 0.17 & 0.08 & 0.96 & 0.48 & 1.54 & 0.78 & 0.88 & 0.44 & \textbf{24.85} & \textbf{13.50} & \textbf{26.82 }& \textbf{14.83 }\\

\midrule
\textbf{LLaMA V1}& 7B & 1 & \textbf{47.27} & \textbf{30.68} & 57.56 & 40.41 & 42.02 & 26.22 & 76.32 & 62.36 & 84.92 & 74.87 & 84.92 & 74.87 &\textbf{ 6.51} & 2.44 & 0.08 & 0.04 & 55.89 & 38.74 \\
\textbf{LLaMA V2}& 7B & 1 & 38.90 & 23.71 & 41.35 & 25.68 & 61.82 & 44.87 & 43.48 & 27.44 & 67.27 & 50.99 & 65.18 & 48.58 & 2.63 & 0.37 & 13.29 & 6.28 & 44.73 & 28.49 \\
\textbf{LLaMA V2}& 13B & 1 & 41.51 & 25.80 & 51.86 & 34.85 & \textbf{75.58} & \textbf{61.38} & 66.42 & 50.00 & 80.06 & 67.58 & 79.34 & 66.55 & 2.83 & 0.47 & 36.26 & 21.65 & 58.18 & 41.04 \\
\textbf{Mistral}& 7B & 1 & 39.32 & 24.04 & 55.80 & 38.65 & 75.44 & 61.19 & 68.70 & 52.68 & 87.53 & 79.04 & 81.22 & 69.25 & 5.62& \textbf{1.96} & \textbf{36.29} & \textbf{21.67} & \textbf{60.60} & \textbf{43.56} \\
\textbf{Mixtral}& 8x7B & 1 & 46.86 & 30.33 & \textbf{58.07} & \textbf{40.93} & 34.54 & 20.35 & \textbf{77.36} & \textbf{63.79} & \textbf{92.33} & \textbf{87.27} & \textbf{92.39} & \textbf{87.38} & 5.31 & 1.79 & 0.77 & 0.40 & 58.54 & 41.41 \\

\midrule
\textbf{LLaMA V1}& 7B & 2 & 42.75 & 26.82 & 71.73 & 56.39 & 79.40 & 66.63 & 82.29 & 70.84 & 87.50 & 78.99 & 87.42 & 78.87 & 41.58 & 25.86 & 0.0 & 0.0 & 66.68 & 50.30 \\

\textbf{LLaMA V2}&  7B & 2 &  52.17 & 35.14 & 72.19 & 56.97  &\textbf{ 93.42} & \textbf{89.24} & 80.64 & 68.41 & 94.05 & 90.41 & 94.16  &  90.60 &  53.40 &  36.31 &  50.95 & 34.01 & 76.52  & 62.64 \\

\textbf{LLaMA V2}& 13B & 2 & 51.22 & 34.26 & \textbf{74.33} & \textbf{59.72} & 93.06 & 88.59  & \textbf{94.01} & \textbf{90.33} & \textbf{95.16} & \textbf{92.48} & \textbf{95.07}  & \textbf{92.32} &  13.53 & 6.42 &  45.55 & 29.19 & 75.80  & 61.66 \\

\textbf{Mistral}& 7B & 2 & 49.33 & 32.52 & 73.29 & 58.38 & 91.48 & 85.76 & 89.79 & 82.83  &  94.58 & 91.40 &  94.74 & 91.69 &  41.66  & 25.92 &  47.71 & 31.08 & 76.38  & 62.45 \\

\textbf{Mixtral}& 8x7B & 2 &43.35  & 27.33 & 63.07 & 46.22 &  78.68 & 65.63 &  69.60 & 53.77 & 84.00 & 73.44 &  83.86 & 73.23 & 10.12  & 4.45 & 27.41  & 15.24 & 61.86  & 44.91 \\

\textbf{Falcon}& 7B & 2 & 47.95 & 31.29 & 65.01 & 48.38 & 81.85 & 70.19 & 89.78 & 82.81  & 85.98 & 76.54 & 84.93  & 74.89 & 13.78  & 6.56 & 0.12  & 0.06 & 65.19  & 48.59 \\

\textbf{Zephyr}& 7B & 2 & \textbf{52.43} & \textbf{35.39} & 71.31 & 55.86 & 92.57 & 87.71 & 90.87 & 84.69 & 94.95 & 92.09 &  94.90 & 91.98 &  \textbf{57.36} & \textbf{40.21} & \textbf{63.35 } & \textbf{46.54} & \textbf{79.52} & \textbf{66.81} \\

\midrule
\textbf{LLaMA V1}& 7B & 3 & 44.45 & 28.25 & 64.32 & 47.61 & 83.55 & 72.75 & 90.72 & 84.44 & 89.96 & 83.12 & 89.76 & 82.78 & 37.07 & 22.28 & 0 & 0 & 68.47 & 52.40 \\

\textbf{LLaMA V2}& 7B & 3 & 52.43 & 35.39 & 71.31 & 55.86 & 92.57 & 87.71 & 90.87 & 84.69 & 94.95 & 92.09 & 94.90 & \textbf{93.98} & \textbf{57.36} & \textbf{40.21} & 63.35 & 46.54 & 79.52 & 66.81 \\

\textbf{LLaMA V2}& 13B & 3 & 49.49 & 32.67 & 66.53 & 50.12 &\textbf{ 93.71} & \textbf{89.78} & \textbf{96.19} & \textbf{94.45} & \textbf{95.17} & \textbf{92.51} & \textbf{95.12} & 92.40 & 41.07 & 25.44 & 53.69 & 36.59 & 77.69 & 64.25 \\

\textbf{Mistral}& 7B & 3 & 51.88 & 34.87 & \textbf{76.11} & \textbf{62.08} & 92.05 & 86.77 & 94.59 & 91.42 & 95.08 & 92.34 & 94.88 & 91.96 & 49.90 & 33.05 & \textbf{66.85} & \textbf{50.50} & \textbf{80.26} & \textbf{67.87} \\

\textbf{Mixtral}& 8x7B & 3 & 51.65 & 34.66 & 69.71 & 53.89 & 87.24 & 78.57 & 80.13 & 67.68 & 93.28 & 88.99 & 93.15 & 88.76 & 20.40 & 10.61 & 29.61 & 16.77 & 70.60 & 54.99 \\

\textbf{Falcon}& 7B & 3 & \textbf{62.30} & \textbf{45.39} & 63.68 & 46.90 & 91.73 & 86.20 & 95.79 & 93.68 & 93.34 & 89.09 & 93.31 & 89.05 & 14.88 & 7.21 & 0.0 & 0.0 & 72.16 & 56.94 \\

\textbf{Zephyr}& 7B & 3 & 56.36 & 39.21 & 72.24 & 57.04 & 84.20 & 73.75 & 87.64 & 79.22 & 93.93 & 90.18 & 93.86 & 90.05 & 57.22 & 40.06 & 51.67 & 34.68 & 76.80 & 63.02 \\

\bottomrule
\end{tabular}
\end{adjustbox}
\label{tab:few_results}
\end{table*}

\begin{table}[h]
\centering
\footnotesize
\begin{minipage}[t]{0.4\textwidth}
\caption{Performance Comparison of Object Classification Methods Across Different Backbone Architectures.}
\label{tab:classification_result}
\centering
\begin{tabular}{@{}lccc@{}}
\toprule
\textbf{Method} & \textbf{ResNet50} & \textbf{VGG16} & \textbf{InceptionV2} \\
\midrule
CAM & 43.10 & \textbf{43.74} & 38.46 \\
HAS & 34.15 & 16.36 & 40.65 \\
ADL & 41.72 & 41.56 & 42.09 \\
ACOL &\textbf{ 43.74 }& 16.36 & 42.09 \\
SPG & 42.19 & - & \textbf{40.92} \\
Cutmix & 41.45 & 41.93 & 40.60 \\
\bottomrule
\end{tabular}
\end{minipage}%
\end{table}

\subsection{Object Detection Results}

We evaluated multiple object detection models including Class Activation Mapping (CAM), Hide-and-Seek (HAS), Attention-based Dropout Layer (ADL), Adversarial Complementary Learning (ACOL), Self-produced Guidance (SPG), CutMix, and DINO across different backbone architectures (ResNet50, VGG16, InceptionV2, Swin Transformer). DINO with Swin 4-scale backbone achieved the best performance with an average score of 32.2 and notable robustness across quality thresholds (45.4 at 10\%). Traditional models like CAM and HAS showed limitations, with ResNet50 scoring averages of 6.17 and 7.14 respectively, with significant performance drops at higher quality thresholds.

Object classification results revealed strong performance from CAM with consistent scores (43.10 for ResNet50, 43.74 for VGG16). HAS showed significant variability, excelling with ResNet50 (34.15) but performing poorly with VGG16 (16.36), indicating dependency on backbone feature extraction capabilities. Advanced methods like ADL and ACOL performed well with InceptionV2 (both achieving 42.09), suggesting that attention diversification across images improves classification in complex datasets.

YOLO model evaluation provided insights into real-time detection capabilities. YOLOv7 demonstrated highest performance with 76.00\% precision, 85.60\% recall, and 79.20\% mAP50, showing robustness in detecting objects on multilingual receipts with diverse layouts. YOLOv8 showed slightly lower performance but improved mAP across different IoU thresholds (45.30\% mAP50-95), suggesting better generalization across object sizes and shapes. YOLOv9 further improved these metrics with 46.70\% mAP50-95, highlighting continuous YOLO architecture advancements.

\subsection{OCR Results}

As shown in Figure~\ref{tab:ocr_performance_comparison}, OCR evaluation focused on Character Error Rate (CER) and Word Error Rate (WER) metrics. Tesseract OCR baseline achieved 15.56\% CER and 30.78\% WER. The Attention-Gated CNN-BiGRU model improved performance with 14.85\% CER and 27.22\% WER by combining gated CNNs with bidirectional GRU layers and attention mechanisms for better spatial dependencies and contextual information handling. Our specialized OCR model reduced CER to 7.83\% but maintained similar WER (27.24\%), indicating better individual character recognition while struggling at word level. Azura OCR achieved best performance with 6.39\% CER and 25.97\% WER.

\subsection{Information Extraction Results}

%Language model evaluation across zero-shot, one-shot, and few-shot scenarios revealed significant variations in structured data extraction from multilingual receipts. 

In zero-shot settings, models performed poorly, particularly for Brand and Weight categories. Zephyr achieved highest performance (42.98 F1, 27.02\% accuracy), while Mistral followed (32.07 F1, 18.53\% accuracy). The \#Units category showed better results with Mistral obtaining 74.94 F1 and 60.52\% accuracy, indicating easier numerical field recognition. One-shot learning showed marked improvement, especially LLaMA V1 in Brand category (jumping from 3.70 to 47.27 F1). Mixtral led Weight extraction (58.07 F1, 40.93\% accuracy), while Mistral achieved best overall performance (60.60 F1, 43.56\% accuracy). Few-shot learning further enhanced performance: in two-shot scenarios, Zephyr recorded 52.43 F1 and 35.39\% accuracy for Brand category, while LLaMA V2 excelled in \#Units extraction (93.42 F1, 89.24\% accuracy). Three-shot setup showed Falcon leading Brand category (62.30 F1, 45.39\% accuracy) and Mistral dominating Weight category (76.11 F1, 62.08\% accuracy), with Mistral achieving best overall performance (80.26 F1, 67.87\% accuracy). 

\subsection{ Model Performance Analysis}
\begin{figure}
    \centering
    \includegraphics[width=0.7\linewidth]{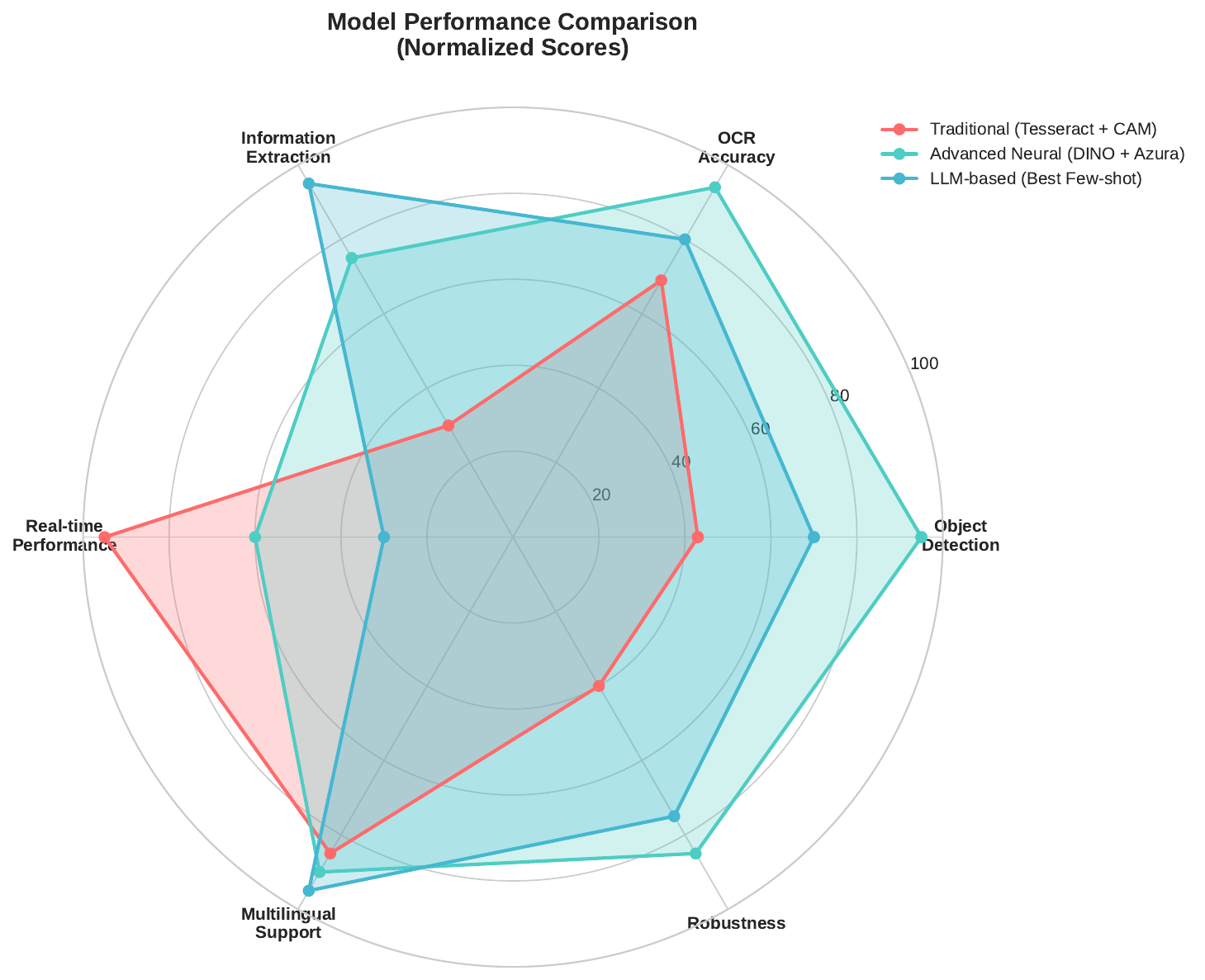}
    \caption{Comprehensive performance comparison of traditional, advanced neural, and LLM-based approaches across six key dimensions for multilingual receipt understanding.}
    \label{fig:model_comparison_radar}
\end{figure}

Figure~\ref{fig:model_comparison_radar} presents a comprehensive comparison of three distinct methodological approaches across six critical performance dimensions for multilingual receipt understanding. The Figure~\ref{fig:model_comparison_radar} reveals complementary strengths and trade-offs between traditional methods, advanced neural networks, and large language models. Traditional approaches (Tesseract + CAM) excel in real-time performance and show reasonable multilingual support but demonstrate significant limitations in object detection, OCR accuracy, and information extraction capabilities. Advanced neural networks (DINO + Azura) achieve superior performance in object detection and OCR accuracy with strong overall robustness, though at the cost of reduced real-time processing speed. LLM-based approaches demonstrate the highest information extraction capabilities and excellent multilingual support but suffer from poor real-time performance due to computational requirements. 

\section{Conclusion }

In this work, we presented \dataset, a comprehensive dataset designed to advance multilingual OCR and post-OCR parsing, particularly for Arabic and English receipts. The dataset consists of over 20,000 annotated receipts, 30,000 OCR-annotated images, and 10,000 item-level annotations, supporting tasks such as object detection, OCR, and information extraction. We evaluated various models, ranging from traditional methods like Tesseract OCR to advanced neural network architectures like YOLO and DINO, as well as large language models like LLaMA and Mistral. Our results demonstrate the dataset's effectiveness in capturing real-world receipt variations and highlight the challenges posed by noisy, multilingual text. By publicly releasing \dataset, we aim to foster research into document understanding across diverse domains and applications.
\section{GenAI Usage Disclosure}
We used OpenAI’s ChatGPT for minor language editing, specifically
to rephrase sentences and correct grammatical errors.

\bibliographystyle{ACM-Reference-Format}
\bibliography{software}

\end{document}